\documentclass[10pt,twocolumn,letterpaper]{article}

\usepackage{cvpr}              % To produce the CAMERA-READY version

\usepackage[utf8]{inputenc} % allow utf-8 input
\usepackage[T1]{fontenc}    % use 8-bit T1 fonts
\usepackage{url}            % simple URL typesetting
\usepackage{booktabs}       % professional-quality tables
\usepackage{amsfonts}       % blackboard math symbols
\usepackage{nicefrac}       % compact symbols for 1/2, etc.
\usepackage{microtype}      % microtypography
\usepackage{bbm}

\usepackage{graphicx}
\usepackage{amsmath}
\usepackage{amssymb}
\usepackage{multirow}
\usepackage{booktabs}
\usepackage{tabularx}
\usepackage[dvipsnames,table,xcdraw]{xcolor}
\usepackage{overpic}
\usepackage{colortbl}
\usepackage{makecell}
\usepackage{contour}

\usepackage[pagebackref=false,breaklinks,colorlinks,citecolor=RoyalBlue,bookmarks=false]{hyperref}

\usepackage{enumitem}
\usepackage{pifont}
\newcommand{\cmark}{\ding{52}}%

\newcommand{\myParaP}[1]{\vspace{.05in}\noindent\textbf{#1}}
\newcommand{\myPara}[1]{\vspace{.05in}\noindent\textbf{#1:}}

\newcommand{\figref}[1]{Fig.~\ref{#1}}
\newcommand{\tabref}[1]{Tab.~\ref{#1}}

\newcommand{\secref}[1]{Sec.~\ref{#1}}

\def\paperID{} % *** Enter the CVPR Paper ID here
\def\confName{CVPR}
\def\confYear{2024}

\graphicspath{{../figure/}, {../figure/Related/}}

\begin{document}

%%%%%%%%% TITLE
\title{Empowering Segmentation Ability to Multi-modal Large Language Models}

\author{
  Yuqi Yang$^{}$\thanks{The first three authors contributed equally to this paper. Peng-Tao Jiang is the corresponding author. } ~~ Peng-Tao Jiang$^{{*}}$ ~~ Jing Wang$^{*}$  ~~ Hao Zhang ~~ Kai Zhao ~~ Jinwei Chen ~~ Bo Li \\
vivo Mobile Communication Co., Ltd. \\
}

\maketitle
%\thispagestyle{empty}

%%%%%%%%% ABSTRACT
\begin{abstract}
Multi-modal large language models (MLLMs) can understand image-language prompts 
and demonstrate impressive reasoning ability.
In this paper, we extend MLLMs' output by empowering MLLMs with the segmentation ability.
The extended MLLMs can both output language responses to the image-language prompts and 
segment the regions that the complex question or query in the language prompts focuses on.
To this end, the existing work, LISA, enlarges the original word embeddings with 
an additional segment token and fine-tunes dialogue generation and query-focused segmentation together, 
where the feature of the segment token is used to prompt the segment-anything model. 
Although they achieve superior segmentation performance, we observe that 
the dialogue ability decreases by a large margin compared to the original MLLMs.
To maintain the original MLLMs' dialogue ability, we propose a novel MLLMs framework, 
coined as LLaVASeg, which leverages a chain-of-thought prompting strategy to 
instruct the MLLMs to segment the target region queried by the user.
The MLLMs are first prompted to reason about the simple description of the target region 
from the complicated user query, then extract the visual attributes of the target region 
according to the understanding of MLLMs to the image.
These visual attributes, such as color and relative locations, 
are utilized to prompt the downstream segmentation model.
Experiments show that the proposed method keeps the original dialogue ability 
and equips the MLLMs' model with strong reasoning segmentation ability.
The code is available at \url{https://github.com/YuqiYang213/LLaVASeg}.
\end{abstract}

%%%%%%%,%% BODY TEXT
\section{Introduction} \label{sec:intro}
Large Language Models (LLMs) show robust dialogue and reasoning skills 
when scaling up the data and model size in Natural Language Processing (NLP).
There emerge massive chatbots based on LLMs~\cite{kenton2019bert,chowdhery2022palm,touvron2023llama,touvron2023llama2}, 
such as OpenAI's chatGPT~\cite{chatGPT}, 
which take language prompts as input and output language responses. 
Since the human perception system can process multiple modal information, 
such as visual and audio information, LLMs hold the potential to be 
extended to multi-modal large language models with multi-modal input.
Thus, prompted by the popular LLMs, Multi-modal Large Language Models (MLLMs) 
have also been widely investigated, which receive not only language 
but also images or audio as input.
There also emerge many extraordinary MLLMs, such as GPT-4V~\cite{gpt4v}, 
Flamingo~\cite{alayrac2022flamingo}, BLIP~\cite{li2023blip}, LLaVA~\cite{chen2023llava}, 
and PaLM~\cite{chowdhery2022palm}, etc.

\begin{figure}[t]
    \centering
    \setlength\tabcolsep{1pt}
    \begin{overpic}[width=0.45\textwidth]{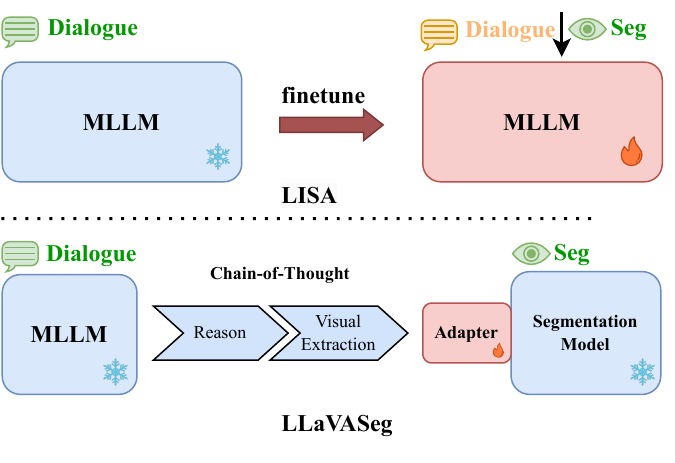}
    \end{overpic}
    \caption{Comparison of previous fine-tuning method \cite{lai2023lisa} and our method. Though previous works empowered the MLLMs with segmentation ability, they downgraded the MLLMs' original reasoning ability. 
    On the contrary, our method not only equips the MLLMs with segmentation ability but maintains the MLLMs' original reasoning ability by freezing parameters.
    }\label{fig:illustration}
\end{figure}

MLLMs act like a person, able to read, see, listen, and then 
generate language responses.
Associating with the human perception system, humans can not only 
understand multi-modal information but also locate the language-focused 
target objects or areas quickly in the physical world.
Thus, some pioneer works~\cite{pi2023detgpt,lai2023lisa,wang2023visionllm} 
extend the capability of MLLMs to different visual tasks with the help 
of instruction fine-tuning.
For instance, Shikra~\cite{chen2023shikra}
encodes the coordinates of the target areas into language and uses them 
to construct instruction-following data with the target location in the output.
Recently, LISA~\cite{lai2023lisa} enlarges MLLMs' word embeddings 
with a $\mbox{<SEG>}$ token.
The output feature of the <SEG> token will prompt the decoder of the 
segment anything model (SAM)~\cite{kirillov2023segment} to segment target areas.
However, we observe that during fine-tuning MLLMs, the MLLM's dialogue ability 
is heavily affected compared with the original MLLMs.
Although LISA introduced instructions from VQA datasets to 
maintain the dialogue ability, 
it cannot completely address this problem and the degradation 
is still significant.
In addition, when prompting the LISA model to output both the segmentation mask and 
the explanations, the quality of the segmentation mask also decreased by a large margin.

In this paper, we aim to empower the MLLMs with segmentation abilities without 
damaging their original reasoning ability.
To this end, one straightforward solution is to prompt the MLLMs to output the name 
of the target area. 
The segmentation model takes the output name and image as input and segments the target 
referred to by its name.
However, since the open-world target name usually doesn't contain any visual information, 
it is challenging for the segmentation model to directly align the target name with a specific image region~\cite{zhu2023llafs}.
Furthermore, this solution does not fully utilize the rich knowledge of the MLLMs 
to facilitate the model locating the target more precisely.

To address this issue, we prompt the MLLMs to generate both the abstract 
target name and detailed image-specific visual attributes.
The visual attributes include the shape, the color, and the relative location of 
the target.
Since the visual attributes can provide further visual information than the target name, 
it can help the segmentation model target objects and refine the segmentation mask.
To generate the visual attributes of the target object from the complicated user query, we propose 
a chain-of-thought prompting to explicitly generate the detailed visual attributes, 
which can be better understood by the segmentation model.
The chain-of-thought prompting includes three steps.
In the first step, we prompt the MLLMs to generate responses 
for understanding the user's query and finding the target from the image.
In this step, the MLLMs tend to generate responses containing lengthy explanations of 
the reasoning step.
In the second step, the MLLMs are prompted to extract the target name 
from lengthy explanations and describe it in the shortest possible way.
In the third step, we prompt the MLLMs to generate visual attributes (color, shape, \etc) 
of the target object and its relative location in the image.
These discriminative attributes are sent to the downstream segmentation model and 
instruct the segmentation model to generate the final segmentation mask.
We propose multi-scale adapters in the segmentation model and fine-tune them to fuse the extracted attributes with the visual features.

Following the above chain-of-thought prompting paradigm, our method, termed LLaVASeg, 
can empower the off-the-shelf MLLMs, such as \textbf{LLaVA},  with \textbf{Seg}mentation ability while not damaging their original reasoning ability. 
We show the comparison of our LLaVASeg and fine-tuning-based method 
in \figref{fig:illustration}.
Experiments show that LLaVASeg achieves superior performance 
in both segmentation and dialogue, which proves the effectiveness of our method.
Unlike LISA~\cite{lai2023lisa} utilize instruction fine-tuning paradigm, 
our LLaVASeg uses chain-of-thought prompting to prompt the off-the-shelf MLLMs 
and performs comparable segmentation performance.
We hope our work can show more insights into the rich capabilities of MLLMs.

In summary, the contributions of this paper are two-fold:
\begin{itemize}
    \item We propose a novel framework that not only equips the MLLMs with 
    segmentation ability but also maintains the MLLMs' original powerful conversational ability. 
    Our method is MLLMs-agnostic and the plug-and-play 
    advantage supports its application to different MLLMs.
    \vspace{2pt}
    
    \item We propose a novel chain-of-thought prompting strategy that 
    iteratively prompts MLLMs to generate suitable responses for downstream segmentation models.
    The chain-of-thought prompting strategy bridges the gap between reasoning and segmentation 
    without additional MLLMs fine-tuning.
\end{itemize}

\section{Related Work} \label{sec:related}
\subsection{Multi-Modal Large Language Model} 
The large language models ~\cite{touvron2023llama,touvron2023llama2,kenton2019bert,chiang2023vicuna,ouyang2022training,diao2023lmflow} 
have exhibited impressive language understanding ability. 
To extend this ability to the domain of vision, many works~\cite{alayrac2022flamingo, li2023blip, chen2023llava, zhu2023minigpt, gpt4v} introduce auxiliary vision models 
to deal with additional visual input.
For example, LLaVA~\cite{liu2023visual}, based on LLaMA~\cite{touvron2023llama}, aligns the visual features 
and language features with instruction tuning.
Recently, some work~\cite{pi2023detgpt, lai2023lisa, liu2023interngpt, chen2023shikra} attempted to enable MLLMs 
with more vision tasks like detection and segmentation.
Shikra~\cite{chen2023shikra} fine-tunes MLLMs on specific instruction-following conversations of which all the 
specific objects are equipped with the coordinates of 
the bounding boxes. 
LISA~\cite{lai2023lisa} broadens the word embedding 
with an additional 
$\mbox{<SEG>}$ token and fine-tunes the multi-modal large language models on 
the instruction-following segmentation data and LLaVA's instruction tuning data.
Different from the above methods, our method does not 
fine-tune the MLLMs and utilizes the chain-of-thought prompting strategy to extract rich information about 
the query-focused region.
The rich information is used to prompt the downstream 
segmentation models.

\subsection{Chain-of-Thought}
LLMs have shown impressive reasoning ability.
Recently, some methods~\cite{wei2022chain, kojima2022large} tend to improve the reasoning ability by prompting LLMs to think step by step.
This prompting strategy, termed chain-of-thought prompting, further 
explores the potential of LLMs.
Boosting with chain-of-thought prompting, LLMs can optimize the reasoning process 
by decomposing the hard task explicitly~\cite{zhou2022least, khot2022decomposed, press2022measuring} or calibrating the results with different chain-of-thoughts~\cite{fu2022complexity, wang2022self}.
Some methods~\cite{tang2023cotdet, zheng2023ddcot, zhang2023multimodal} extend 
the chain-of-thought prompting to the vision domain.
Among them, CoTDeT~\cite{tang2023cotdet} leverages the chain-of-thought prompting to explore the affordance required by specific tasks to benefit object detection.
Different from these methods, our method proposes to leverage the chain-of-thought 
prompting to prompt the MLLMs to generate the visual attributes, such as the shape, the color and the
relative location of the query-focused image region.
The visual attributes can provide rich visual information about the target, 
facilitating the segmentation model to locate target regions and refine the segmentation mask.

\begin{figure*}[t]
    \centering
    \setlength\tabcolsep{1pt}
    \begin{overpic}[width=\textwidth]{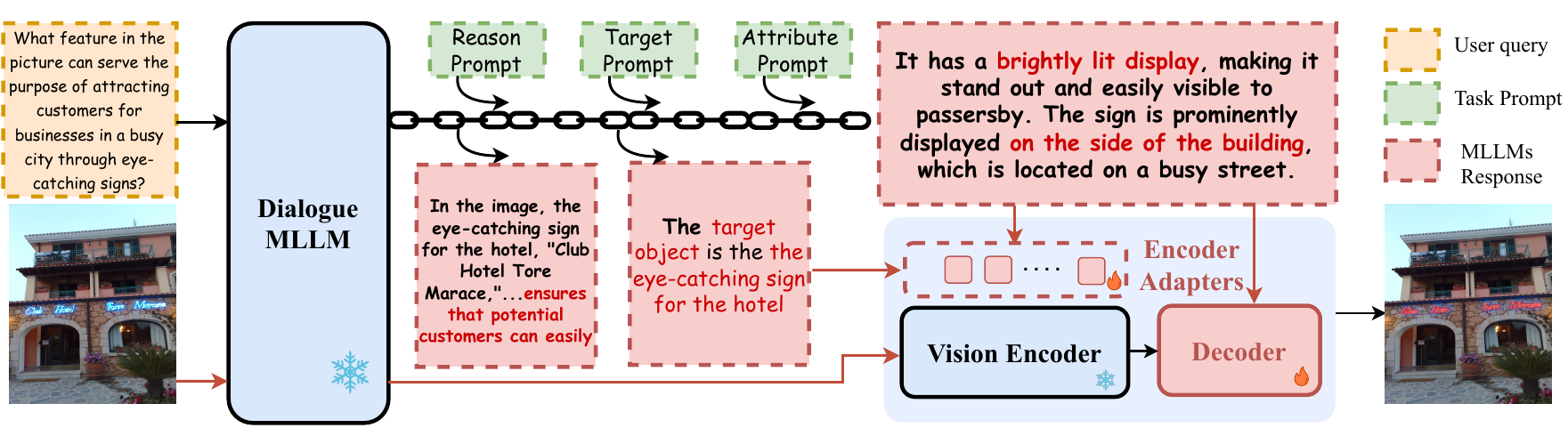}
    \end{overpic}
    \caption{Overall pipeline of our LLaVASeg. 
    Given the user query and image, the MLLMs generate the visual attributes through chain-of-thought prompting.
    Then, we input these visual attributes into the segmentation model to perform multi-scale prompting and generate the segmentation mask.
    During training, the MLLMs and the segmentation model are frozen, and only the lightweight adapters are trainable to guarantee efficiency.
    }\label{fig:pipeline}
\end{figure*}

\subsection{Referring Segmentation} 
The referring segmentation aims to segment the objects or regions 
pointed out by the text description.
A typical paradigm for referring segmentation methods~\cite{ding2021vision, feng2021encoder, yang2022lavt} is combining the text features into the 
visual features to assist in segmenting the target objects.
Specifically, VLT~\cite{ding2021vision} and EFN ~\cite{feng2021encoder} fuse visual and text features in the decoder stage, while LAVT ~\cite{yang2022lavt} choose to combine the text and visual features at the encoder stage and obtain better performance.
However, most existing referring segmentation datasets~\cite{kazemzadeh2014referitgame} 
rely on a simple description, resulting in many referring segmentation methods 
lacking reasoning and analyzing ability.
Our method, with the help of MLLMs, performs a better understanding of complex queries 
that require common knowledge and understanding of the world.

\section{Method}
In this section, we will introduce our method in detail.
The framework of the proposed LLaVASeg is shown in \figref{fig:pipeline}.
First, we will introduce the problem definition and the pipeline of MLLMs 
in \secref{sec:preliminaries}.
Subsequently, we will dive into the details of our chain-of-thought prompting 
design in \secref{sec:cot}.
Then, we will introduce the prompting segmentation network in \secref{sec:seg}.
Finally, we discuss our data construction pipeline in \secref{sec:data}.

\subsection{Preliminaries}
\label{sec:preliminaries}
\myParaP{Reasoning Segmentation} is first proposed in ~\cite{lai2023lisa}, 
which aims to parse the question-oriented or query-focused visual region 
in an image and output a segmentation mask to locate it.
Although this task shares similarities with referring segmentation task~\cite{kazemzadeh2014referitgame}, it requires more knowledge 
about the real world and common sense to reason about the target region 
from intricate expression.

\myParaP{The Pipeline of MLLMs}
The MLLMs take multi-modal information as input that includes 
an image $\mathbf{I} \in \mathcal{R}^{H \times W \times 3}$ 
and text $\mathcal{T}$ of random length.
The image $\mathbf{I}$ is encoded with the multi-modal encoder $v$ to share 
the same embedding space with the text embedding~\cite{chen2023llava}.
The text can be further divided into two parts.
The first part is the user query $\mathcal{T}_{q}$, which is related 
to the requirements of the user.
The second part is the pre-designed task prompt $\mathcal{T}_{p}$, 
which is hand-crafted or generated automatically.
Although the prompt is not given by the user, 
a proper prompt can instruct the MLLMs to generate a response 
with higher quality~\cite{kojima2022large}.
Formally, the MLLMs $f_{l}$ generate the answer $\mathcal{T}_{a}$ by
\begin{equation}
    \mathcal{T}_{a} = f_{l}(v(\mathbf{I}), \{\mathcal{T}_{p}, \mathcal{T}_{q}\}),
\end{equation}
where $\{\cdot\}$ indicates the concatenation of text input.
The MLLMs also support multi-turn conversation by concatenating the former input 
and response with another query and prompt, which can be formulated as:
\begin{equation}
    \mathcal{T}^{\prime}_{a} = f_{l}(v(\mathbf{I}), \{\{\mathcal{T}_{p_1}, \mathcal{T}_{q_1}, \mathcal{T}_{a}\},\mathcal{T}_{p_2},\mathcal{T}_{q_2}\}),
\end{equation}
where $\mathcal{T}_{p_i}$ and $\mathcal{T}_{q_i}$ indicate the task prompt and user query in the $i$-th turn.

Recently, LISA~\cite{lai2023lisa} adapted the MLLMs to the segmentation task 
by enlarging the word embeddings with a <SEG> token.
The features of the <SEG> token are used to prompt the downstream segmentation model.
However, the instruction tuning for the segmentation task heavily affects the conversational and reasoning ability of the original MLLMs.
In this paper, instead of fine-tuning the MLLMs, we attempt to change 
the prompt $\mathcal{T}_{p}$ to generate responses containing the name and 
the visual attributes of the target.
The visual attributes include the shape, color and relative location of the target, 
which provides additional image-specific information for the target.
The prompt-based method can maintain the conversation and reasoning ability of MLLMs.

\subsection{Chain-of-Thought Prompting Design}
\label{sec:cot}
To generate the visual attributes, knowledge from both the real world 
and the given image is needed to link the complicated user query 
to the target area in the image.
When naively prompting the MLLMs to reason from the user query, 
the MLLMs may only generate the rationales or explanations in response to the query, 
not the needed visual attributes.
To address this problem, we propose to leverage chain-of-thought prompting, 
which explicitly prompts MLLMs to generate the visual attributes of 
query-focused regions step by step. 
Our chain-of-thought prompting includes three steps: reason prompting step, 
target prompting step and attribute prompting step.
In the first step, we prompt the MLLMs to use common knowledge to find the target area.
Since the response in the first step usually includes many unrelated explanations, we then prompt the MLLMs to find the exact target name and get rid of unrelated information in the second step.
In the third step, we prompt the MLLMs to extract the visual attributes 
of the target area from the image.
We will dive into the details of these steps in the rest of this section.

\myPara{Reason Prompting Step}
The first step is to prompt MLLMs to reason about the target region 
in the image from the user query.
The reasoning step can be formulated as a VQA task, so we rephrase the user query 
into a question for a specific object or area in this image.
Concretely, we design the following prompt for this step:

\vspace{0.4em}
\noindent\textit{Prompt: What is the object or part that is \textcolor{black}{[USER QUERY]} in this image?}

\noindent\textit{Output: It is fire in the fire pit. The fire is hot and ...}
\vspace{0.4em}

\noindent [USER QUERY] is filled with the user query.
When the user query is a question, we directly use it as the input and do not 
add other prompts.
Notice that we do not instruct the MLLMs to output the rationale explicitly.
We find that the MLLMs tend to answer the question followed by the explanation, 
although no explicit prompt is given.
Additionally, forcing the model to output the rationale while not providing 
any knowledge for it may result in severe hallucination~\cite{zhang2023multimodal}.
Given these concerns above, we leave the reason prompt as a question and 
do not prompt the MLLMs to give a detailed explanation explicitly.

\myPara{Target Prompting Step}
The MLLMs output both the name of the target region and the explanation including many 
unrelated descriptions (e.g. hot, light...) in the first step.
These descriptions may disturb the MLLMs to generate exact visual attributions in the attribute prompting step, and it is not trivial to find the exact target from the response.
In the second step, we input the user query, response, 
and image to the MLLMs to provide sufficient information 
so that the MLLMs can analyze the target.
Specifically, the target prompt is formulated as follows:

\vspace{0.4em}
\noindent\textit{Prompt: Here is the conversation:}

\noindent\textit{The question is: [QUESTION]}

\noindent\textit{The answer is: [ANSWER]}

\noindent\textit{Please analyze the conversation and identify the distinct 
physical objects or areas that the user wants from the image. }

\noindent\textit{Output: The user wants the fire from the image.}
\vspace{0.4em}

\noindent [QUESTION] and [ANSWER] is the question and answer in the first step, respectively.

\myPara{Attribute Prompting Step}
In this step, we prompt the MLLMs to provide visual information towards the target.
Although it seems straightforward to prompt the MLLMs to output the location of the target object 
in the image, previous work~\cite{chen2023shikra} shows that 
the MLLMs cannot output accurate spatial positions in zero-shot settings.
To address this issue, we choose to extract the visual attributes rather than the specific location 
to aid the downstream segmentation model.
The visual attributes include color, shape, and the relative position to the other objects 
in the image.
We prompt the MLLMs to extract the visual attributes as follows:

\vspace{0.4em}
\noindent\textit{Prompt: Here is the target:}

\noindent\textit{The target is: [TARGET]}

\noindent\textit{Briefly describe the target entity or part's visual attributes that can discriminate them from the image.
Each visual attribute can be color, shape, and relative position to other objects in the image. }

\noindent\textit{Output: The fire can be discriminated from the image by its bright orange color and the fact that it is emitting heat and light. The fire is surrounded by ...}
\vspace{0.4em}

\noindent [TARGET] is the target analyzed in the second step.
For simplicity, we present the most critical parts of the prompts.
Note that we prompt the MLLMs to generate the discriminative visual attributes 
which are more helpful in segmenting the target region 
from the image.

In the implementation, the target prompting step and the attribute prompting step are performed in the same round of conversation for efficiency.
Specifically, we prompt the MLLMs to first target the region and then extract its visual attributes explicitly as follows: \textit{Follow these guidelines strictly: (1) Please analyze the conversation... (2) Briefly describe the target entity or...}.
We empirically find that this can save inference time 
without an apparent performance drop.

\subsection{Multi-Scale Prompting Segmentation}
\label{sec:seg}

Since the extracted visual attributes can serve in both low-level (e.g. color) and high-level (e.g. name, relative location), we propose a multi-scale promptable segmentation pipeline to align the visual attributes and their corresponding area.
In this section, we introduce the architecture and 
the training objectives of our segmentation pipeline.

\myPara{Segmentation Pipeline}
We input the target name with the visual attributes $\mathbf{E}$ and the image $\mathbf{I}$ into 
the segmentation model \(f_s\) to generate the final mask \(\mathbf{M}\), which 
can be formulated as:
\begin{align}
    \mathbf{M} &= f_s(\mathbf{I}, \mathbf{E}).
\end{align}
In our method, we directly use MLLMs' token 
embeddings corresponding to the target name and 
visual attributes and don't utilize extra 
text encoders like BERT~\cite{kenton2019bert}.
We add an MLP to project these textual embeddings 
to the visual feature space.
Given the projected textual embeddings $\mathbf{E}$, to perform multi-scale interaction between $\mathbf{E}$ and the visual features in the segmentation model, we introduce several language-aware adapters in different layers.
Motivated by~\cite{yang2022lavt}, we design our language-aware module based on the attention mechanism.
Specifically, we leverage the cross-attention between the textual embedding and the visual features from the segmentation model to inject the visual attributes into the segmentation model.

Formally, we pick up several layers in the segmentation backbone to 
set the language-aware modules.
When the language-aware module is set in $p$-th layer, the visual feature $\mathbf{V}_{p}$ will be fed into the language-aware module together with language embedding $\mathbf{E}$ to generate the fused feature $\mathbf{F}_{p}$, which can be formulated as:
\begin{align}
    \mathbf{F}_{p} &= \mathcal{LAM}_{p}(\mathbf{V}_{p}, \mathbf{E}).
\end{align}
Each $\mathcal{LAM}_{p}$ mainly consists of a cross-attention module 
and a feed-forward layer.
It takes $\mathbf{V}_{p}$ as query and $\mathbf{E}$ as key and value.
After the interaction, the fused feature will be added by the original features $\mathbf{V}_p$ through a residual path.

For further interaction between visual features and textual features $\mathbf{E}$ 
in the decoding phrase, we take inspiration from SAM~\cite{kirillov2023segment} 
and leverage the prompt-based decoder in our method.
The decoder uses cross-attention in both directions, which is prompt-to-vision embedding and vice-versa, as its main components.

\myPara{Training Objectives}\label{sec:loss}
During training, we freeze all the parameters of the segmentation model 
except for the adapter and decoder.
Different from LISA~\cite{lai2023lisa}, our model is only trained with the segmentation 
loss $\mathcal{L}_{mask}$. 
Specifically, our loss $\mathcal{L}_{mask}$ is a combination of per-pixel binary cross-entropy (BCE) loss and DICE loss, which can be formulated as 
\begin{align}
    \mathcal{L}_{mask} = \lambda_{bce} \textbf{BCE}(\hat{\mathbf{M}}, \mathbf{M}) + \lambda_{dice} \textbf{DICE} (\hat{\mathbf{M}}, \mathbf{M}),
\end{align}
where $\hat{\mathbf{M}}$ is the ground-truth and $\mathbf{M}$ is the prediction.
$\lambda_{bce}$ and $\lambda_{dice}$ denote the loss weights.

\subsection{Training Data Construction}\label{sec:data}
To train LLaVASeg, we collect two kinds of training data. 
The first one is the segmentation dataset with complicated user queries, including ReasonSeg~\cite{lai2023lisa}.
We also use its validation set to evaluate the performance of our method.
However, ReasonSeg only contains 1218 image-instruction pairs.
The limited data is not enough to align the feature from MLLMs and the segmentation model, 
and will heavily harm the generalization ability of our LLaVASeg.
To address this problem, we collect the segmentation dataset without complicated queries as our second kind of data.
It includes referred segmentation datasets~\cite{kazemzadeh2014referitgame, mao2016generation} and semantic segmentation datasets~\cite{zhou2017scene, caesar2018coco}.
Further introduction for these datasets is in \secref{sec:exp}.
In these datasets, only a brief description of the target is provided.
This makes it difficult to construct the full chain-of-thought prompting when training with these datasets.
As a result, we use different Q-A templates to simulate the first step of chain-of-thought prompting.
Since the parameters of MLLMs are frozen throughout the training, this will not affect the quality of reasoning.
A large number of training samples can effectively promote our LLaVASeg with strong segmentation and generalization ability.

\section{Experiment} \label{sec:exp}

\subsection{Experimental Setting}

\myPara{Datasets}
Following LISA~\cite{lai2023lisa}, our method adopts 
a composed dataset containing multiple datasets from different tasks.
These datasets are collected from three categories as mentioned in \secref{sec:data}.
The first category is semantic segmentation datasets, 
including ADE20k~\cite{zhou2017scene}, COCO-Stuff~\cite{caesar2018coco}, and 
part semantic segmentation dataset PACO-LVIS~\cite{ramanathan2023paco}, and PASCAL-Part~\cite{chen2014detect}.
The second is referring segmentation datasets, including RefCOCO~\cite{kazemzadeh2014referitgame}, RefCOCO+~\cite{kazemzadeh2014referitgame}, RefCOCOg~\cite{mao2016generation}, Refclef~\cite{kazemzadeh2014referitgame}.
The datasets mentioned above provide a large number of visual samples with different granularity to promote our LLaVASeg with strong segmentation ability and generalization abilities.
The third category is the reasoning segmentation dataset ReasonSeg~\cite{lai2023lisa}.
We evaluate the performance of the conversational ability 
based on the reasoning segmentation dataset by evaluating 
the MLLMs' response in the reason prompting step.
Furthermore, we exclude the COCO samples in the refCOCO(+/g) validation set to avoid data leakage.

\myPara{Implementation Details}
For all the experiments, we use LLaVA~\cite{liu2023llava} as the architecture of MLLMs if not specified.
The backbone of LLaVA is set as Llama2~\cite{touvron2023llama2}.
For the segmentation model, we choose the backbone of ViT-H SAM~\cite{kirillov2023segment} as the vision backbone.
For model training, AdamW~\cite{loshchilov2017decoupled} 
is used as our optimizer.
The overall learning rate is set as 0.0001, and the 
weight decay is set to 0.0001.
$\lambda_{bce}$ and $\lambda_{dice}$  are set to $1$ and $0.5$, 
respectively.
The overall batch size is set to 160, where each sample 
contains a pair of text and a referred mask.
We train our model for 12000 iterations in total.

\myPara{Evaluation Metrics}
Similar to LISA~\cite{lai2023lisa}, we adopt gIoU (the 
average of all per-image intersection-over-unions) 
and cIoU (the cumulative intersection over
the cumulative union) as evaluation metrics on reasoning segmentation.
We evaluate our framework's dialogue quality based on the 
CIDEr metric~\cite{vedantam2015cider} and ROUGE-L metric~\cite{lin2004rouge}.  
ROUGE-L is a metric commonly used in natural language processing for evaluating the quality of summaries by calculating the length 
of the longest common subsequences between the generated summary and the reference summary.
CIDEr~\cite{vedantam2015cider} is a metric specifically designed for evaluating the quality of image captions. 
It takes into account consensus and diversity among reference captions, 
providing a more comprehensive evaluation of the generated 
image captions.
We adopt these two evaluation metrics for a more complete and comprehensive evaluation of the generated response 
from our framework and LISA.

\begin{table}[t]
    \centering
    \renewcommand{\arraystretch}{1.0}
    \tabcolsep=0.6mm
    {
    \caption{Reasoning segmentation results of LLaVASeg (ours) and previous related works. 'explain' denotes that we use the LISA model that can output both the segmentation mask and its explanation. $\uparrow$ denotes higher is better.}
    \label{table:reason_seg}   
        \begin{tabular}{l|cc|cc}
            \multirow{2}*{Method} & \multicolumn{2}{c|}{ReasonSeg val} & \multicolumn{2}{c}{ReasonSeg val} \\

            \specialrule{0em}{0pt}{1pt}
            \cline{2-5}
            \specialrule{0em}{0pt}{1pt}
            
            ~ & gIoU $\uparrow$ & cIoU $\uparrow$ & $\text{ROUGE-L}$ $\uparrow$ & CIDEr $\uparrow$ \\
            
            \specialrule{0em}{0pt}{1pt}
            \hline
            \specialrule{0em}{0pt}{1pt}

            OVSeg~\citep{liang2023open} & 28.5 & 18.6 & - & - \\ 

            GRES~\citep{liu2023gres} & 22.4 & 19.9 & - & - \\    % 
            
            X-Decoder~\citep{zou2023generalized} & 22.6 & 17.9 & - & - \\

            SEEM~\citep{zou2023segment} & 25.5 & 21.2 & - & - \\

            \specialrule{0em}{0pt}{1pt}
            \hline
            \specialrule{0em}{0pt}{1pt}
            
            LISA-7B & 44.4 & 46.0 & - & - \\
            LISA-13B (explain) & 57.3 & \textbf{60.7} &0.290 & 0.107 \\
            LLaVASeg-7B & 54.8 & 49.9 & - & - \\
            LLaVASeg-13B & \textbf{59.1} & 52.8 & \textbf{0.393} & \textbf{0.796} \\
                 
        \end{tabular}
    }
\end{table}

\begin{figure}[t]
    \centering
    %\footnotesize
    \setlength\tabcolsep{1pt}
    \begin{overpic}[width=0.45\textwidth]{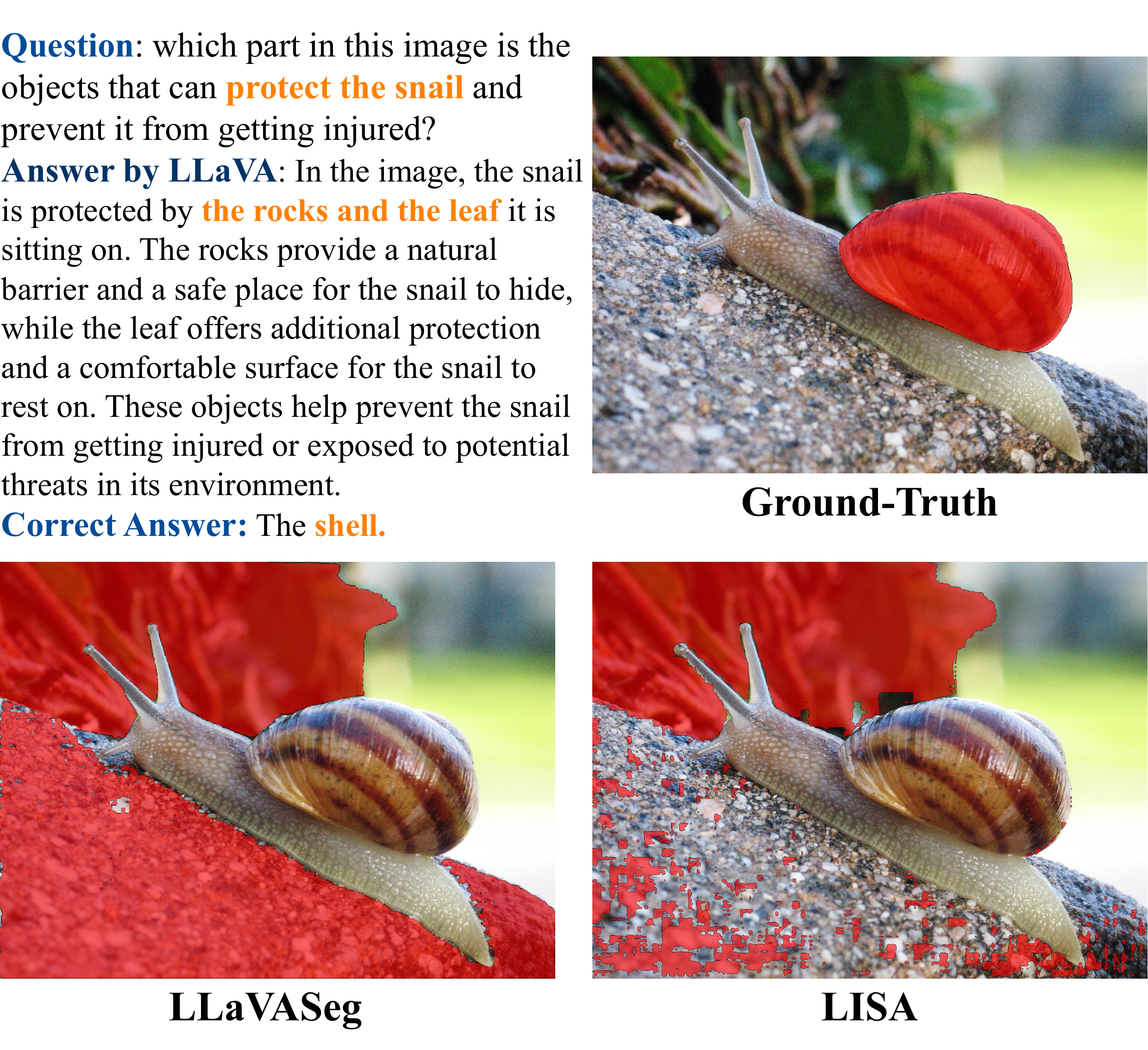}
    % \put(5.5, 30){global view}
    \end{overpic}
    \caption{A failure case study for LLaVASeg and LISA.
    The upper left corner shows that the LLaVA fails to reason the correct answer from the user query. 
    }\label{fig:fail}
\end{figure}

\subsection{Reasoning Segmentation}
\myPara{Segmentation}
We evaluate the segmentation performance of our LLaVASeg 
and compare it with the previous methods on the ReasonSeg dataset. 
For a fair comparison with LISA, we utilize the 
`LISA-13B-explain` model which can output both the explanation 
and the segmentation mask.
The result is shown in \tabref{table:reason_seg}.
It can be seen that our LLaVASeg achieves superior performance on segmentation, 
which is 1.8\% higher than LISA in terms of the gIoU metric.
Furthermore, we should note that our LLaVASeg does not fine-tune LLaVA on 
the Reasonseg dataset, and inherently makes mistakes in reasoning and 
results in lower cIoU than LISA.
We present a failure case study as shown in \figref{fig:fail}.
In this case, when the LLaVA model fails to reason about the correct answer, 
both LLaVASeg and LISA predict the wrong area.
However, due to LLaVASeg's superior segmentation ability, it predicts the area 
referred by the LLaVA response more completely than LISA.
Consequently, such samples result in higher cumulative union 
and lower cIoU.
Besides, we also observe that gIoU is much more stable than cIoU since cIoU is largely biased towards large-area target~\cite{lai2023lisa}.
This experiment shows that our model can achieve competitive segmentation performance without fine-tuning MLLMs.

\begin{figure*}[t]
    \centering
    \setlength\tabcolsep{1pt}
    \begin{overpic}[width=0.97\textwidth]{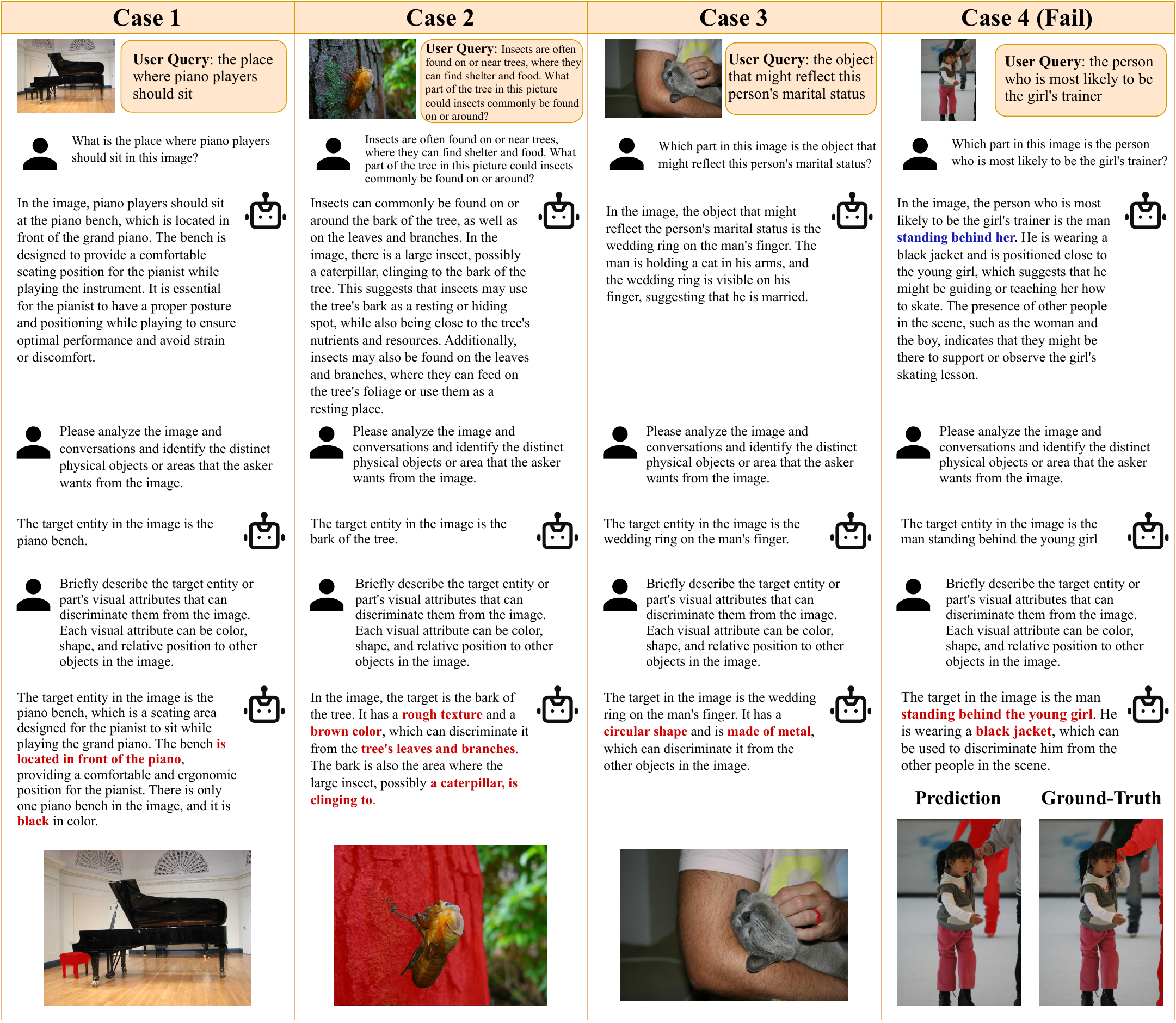}
    \end{overpic}
    \caption{Visualization of the results of our pipeline.
    We also show one failure case caused by a mistake in the reasoning step for a deeper understanding to our method.
    }\label{fig:vis_case}
\end{figure*}

\myPara{Explanation Generation}
To investigate the impact of fine-tuning MLLMs for segmentation tasks on their original 
reasoning ability and dialogue ability, 
we evaluate the explanation performance.
Specifically, we prompt both the LISA and the off-the-shelf LLaVA used in our LLaVASeg, 
to produce the answer and the explanation to the user query.
The experiment's results, presented in \tabref{table:reason_seg}, reveal a notable decline in the dialogue performance when comparing LISA with LLaVA.
These findings support our motivation and prove the significance of our work 
that empowers the MLLMs with segmentation ability while preserving their reasoning ability.
We also present visualization results in \figref{fig:vis_case} 
for an intuitive understanding of our method.

\begin{table*}[htb]
    \centering
    \tabcolsep=0.3cm
    {
    \caption{Results on referring segmentation of LLaVASeg and other existing methods. All the results are evaluated with the cIoU metric.
    }
    \label{table:refer_seg}   
        \begin{tabular}{ c | c c c | c c c | c c }
            
            \multirow{3}*{Method} & \multicolumn{3}{c|}{refCOCO} & \multicolumn{3}{c|}{refCOCO+}  & \multicolumn{2}{c}{refCOCOg} \\ 
            
            \specialrule{0em}{0pt}{1pt}
            \cline{2-9}
            \specialrule{0em}{0pt}{1pt}
            
            ~ & val & testA & testB & val & testA & testB & val(U) & test(U) \\

            \specialrule{0em}{0pt}{1pt}
            \hline
            \specialrule{0em}{0pt}{1pt}

            MCN~\citep{luo2020multi} & 62.4 & 64.2 & 59.7 & 50.6 & 55.0 & 44.7 & 49.2 & 49.4 \\

            VLT~\citep{ding2021vision} & 67.5 & 70.5 & 65.2 & 56.3 & 61.0 & 50.1 & 55.0 & 57.7 \\

            CRIS~\citep{wang2022cris} & 70.5 & 73.2 & 66.1 & 62.3 & 68.1 & 53.7 & 59.9 & 60.4 \\

            LAVT~\citep{yang2022lavt} & 72.7 & 75.8 & 68.8 & 62.1 & 68.4 & 55.1 & 61.2 & 62.1 \\
            
            ReLA~\citep{liu2023gres} & 73.8 & 76.5 & 70.2 & \textbf{66.0} & {71.0} & {57.7} & 65.0 & 66.0 \\
            
            X-Decoder~\citep{zou2023generalized} & - & - & - & - & - & - & 64.6 & -  \\

            SEEM~\citep{zou2023segment} & - & - & - & - & - & - & 65.7 & -    \\

            LISA-7B & 74.9 & 79.1 & 72.3 & 65.1 & 70.8 & \textbf{58.1} & {67.9} & \textbf{70.6} \\

            \specialrule{0em}{0pt}{1pt}
            \hline
            \specialrule{0em}{0pt}{1pt}

            LLaVASeg-7B  & \textbf{76.2} & \textbf{79.1} & \textbf{72.9} & {65.7} & \textbf{71.4} & {57.7} & \textbf{69.8} & {70.1}\\         
        \end{tabular}
    }
\end{table*}

\begin{table}[htb]
\centering
\caption{\textbf{Ablation study.} We perform experiments on ReasonSeg with LLaVSeg. All the experiments are done in LLaVASeg-13B except for (a).}
\subfloat[Ablation on different prompts. 
    The step refers to that described in Sec.\ref{sec:cot}.
    Specifically, Step 1 refers to the reason prompting step. Step 2 refers to the target prompting step. Step 3 refers to the attribute prompting step.\label{table:different_prompts}]{
    \tabcolsep=0.32cm
\begin{tabular}{ccc|cc}
    Step 1 & Step 2 & Step 3 & gIoU & cIoU \\
    \midrule
        \cmark & ~ & ~ & {36.7} & {31.4} \\
        \cmark & \cmark & ~ & {50.2} & {43.8} \\
       \cmark & \cmark & \cmark  & {54.8} & {49.9} \\
\end{tabular}}\hspace{10mm}
\subfloat[Ablation the number of scales when prompting the segmentation model.\label{table:multi_scale}]{
\tabcolsep=0.28cm
\begin{tabular}{c|c|cc}
    Settings & Backbone & gIoU & cIoU \\
    \midrule
    Single-scale & LLaVA-13B & {55.4} & {52.5} \\
    Multi-scale  & LLaVA-13B  & {59.1} & {52.8} \\
\end{tabular}}\hspace{10mm}
\subfloat[Segmentation performance when replacing chain-of-thought with LISA.\label{table:agent_structure}]{
\tabcolsep=0.11cm
\begin{tabular}{c|c|cc}
     {Method} & {Backbone} & gIoU & cIoU \\
     \hline
     Replaced with LISA & LLaVA-13B & {54.3} & {50.9} \\
     LLaVASeg & LLaVA-13B & {59.1} & {52.8} \\
\end{tabular}}\hspace{10mm}
\subfloat[Ablation on generalization ability. 
'nors' denotes that our model has not been trained with the ReasonSeg
dataset. \label{table:vision_ft}]{
\tabcolsep=0.23cm
\begin{tabular}{c|c|cc}
 {Method} & {Backbone} & gIoU & cIoU \\ 
 \hline
 LLaVASeg (nors) & LLaVA-13B & {52.8} & {45.3} \\
 LLaVASeg & LLaVA-13B & {59.1} & {52.8} \\
\end{tabular}}\hspace{10mm}
\end{table}

\myPara{Vanilla Referring Segmentation}
We evaluate our method on the referring segmentation datasets to show that the proposed LLaVASeg framework can also handle the vanilla referring segmentation task.
As shown in \tabref{table:refer_seg}, our method achieves competitive performance compared with previous works, which outperforms the previous works on most of the dataset splits.
This demonstrates the superior segmentation ability of our LLaVASeg.

\subsection{Ablation Study}

\myPara{Different Prompts}
To show the effectiveness of the proposed chain-of-thought prompting, 
we ablate on the performance of our LLaVASeg when using different prompts.
Specifically, we use only one or two steps of our chain-of-thought prompting 
in this ablation for comparison.
The results are shown in \tabref{table:different_prompts}.
It can be seen from the results that when only the first step is leveraged, 
the performance drops significantly.
When the targeting prompting step is added, the performance increases.
This indicates that the unrelated rationale and explanation hinder the segmentation severely.
When we further add the attribute prompting step, the performance 
also increases by a large margin, which demonstrates the effectiveness 
of the visual attributes.
This underscores the effectiveness of our chain-of-thought prompting 
in maximizing the reasoning segmentation accuracy.

\myPara{Effectiveness of Multi-Scale Prompting}
We ablate on the impact of the multi-scale features on 
the reasoning segmentation performance.
The results are shown in \tabref{table:multi_scale}.
It can be seen that the setting using the multi-scale features 
achieves superior performance than using single-scale features.
the performance of both gIoU and cIoU increases accordingly.
These results demonstrate that the extracted vision attributes 
including both low-level and high-level information can benefit 
from the multi-scale features. 

\myPara{Replacing Chain-of-Thought with LISA}
In this experiment, we show the effectiveness of the explicitly generated visual attributes.
We employed LISA, which extracts the visual attributes implicitly in the <SEG> token, to undertake the second and third steps 
in the chain-of-thought prompting as the comparison.
Specifically, we give the MLLMs response from the first step 
to the LISA as the instruction for segmentation.
The results are shown in \tabref{table:agent_structure}.
We can see that LISA achieves a gIoU of 54.3\% and a cIoU of 50.9\%, and LLaVASeg outperforms it with a gIoU of 59.1\% and a cIoU of 52.8\%.
We argue that the significant improvements mainly come from 
the explicit visual attributes extraction in our chain-of-thought 
prompting when comparing with LISA.
With the same reasoning response as input, our LLaVASeg leverages an explicitly generated visual attribute to aid the segmentation model, while LISA does this implicitly in the <SEG> token.

\myPara{Ablation on Generalization Ability}
The experiments are evaluated on the ReasonSeg validation set.
To test the generalization ability of LLaVASeg, we train 
LLaVASeg on the dataset excluding the ReasonSeg training set.
The results in \tabref{table:vision_ft} show that there is 
a performance drop when the ReasonSeg training 
set is excluded.
Although the performance degrades, our LLaVASeg without training 
on the ReasonSeg dataset still achieves competitive performance, which further supports its generalization ability.

\section{Limitations and Future Works}
The main limitations of our methods are two-fold.
Firstly, the current LLaVASeg only supports one query in one round of interaction.
A prompt design supporting multiple queries in one round awaits to be studied.
Secondly, our LLaVASeg leverages off-the-shelf MLLMs. 
However, its performance can be further improved by instruction tuning with high-quality chain-of-thought instruction pairs.
Given that the chain-of-thought instruction pairs are fully text rather than the visual prompt <SEG>, it can be expected that there is less impact on its conversational ability.
To sum up, we aim to develop a more comprehensive prompting design and dive into the potential of instruction tuning in our future research.

\section{Conclusion}
In this work, we present a novel reasoning segmentation framework LLaVASeg. 
LLaVASeg equips the MLLMs with segmentation ability while 
maintaining their conversational and reasoning ability.
To this end, we propose a novel chain-of-thought prompting strategy
that iteratively prompts MLLMs to generate image-specific textual attributes 
for prompt the segmentation model.
To better use these attributes, we leverage a multi-scale prompting segmentation pipeline 
and train lightweight language-aware adapters for the interaction between 
textual attributes and visual features.
Our method performs competitive segmentation ability by using the off-the-shelf MLLMs.
We hope our work will provide valuable inspiration for 
the marrying of MLLMs and different vision tasks.

{\small
\bibliographystyle{ieee_fullname}
\bibliography{llavaseg}

\begin{thebibliography}{10}\itemsep=-1pt

\bibitem{alayrac2022flamingo}
Jean-Baptiste Alayrac, Jeff Donahue, Pauline Luc, Antoine Miech, Iain Barr,
  Yana Hasson, Karel Lenc, Arthur Mensch, Katherine Millican, Malcolm Reynolds,
  et~al.
\newblock Flamingo: a visual language model for few-shot learning.
\newblock {\em Advances in Neural Information Processing Systems},
  35:23716--23736, 2022.

\bibitem{caesar2018coco}
Holger Caesar, Jasper Uijlings, and Vittorio Ferrari.
\newblock Coco-stuff: Thing and stuff classes in context.
\newblock In {\em Proceedings of the IEEE conference on computer vision and
  pattern recognition}, pages 1209--1218, 2018.

\bibitem{chen2023shikra}
Keqin Chen, Zhao Zhang, Weili Zeng, Richong Zhang, Feng Zhu, and Rui Zhao.
\newblock Shikra: Unleashing multimodal llm's referential dialogue magic.
\newblock {\em arXiv preprint arXiv:2306.15195}, 2023.

\bibitem{chen2023llava}
Wei-Ge Chen, Irina Spiridonova, Jianwei Yang, Jianfeng Gao, and Chunyuan Li.
\newblock Llava-interactive: An all-in-one demo for image chat, segmentation,
  generation and editing.
\newblock {\em arXiv preprint arXiv:2311.00571}, 2023.

\bibitem{chen2014detect}
Xianjie Chen, Roozbeh Mottaghi, Xiaobai Liu, Sanja Fidler, Raquel Urtasun, and
  Alan Yuille.
\newblock Detect what you can: Detecting and representing objects using
  holistic models and body parts.
\newblock In {\em Proceedings of the IEEE conference on computer vision and
  pattern recognition}, pages 1971--1978, 2014.

\bibitem{chiang2023vicuna}
Wei-Lin Chiang, Zhuohan Li, Zi Lin, Ying Sheng, Zhanghao Wu, Hao Zhang, Lianmin
  Zheng, Siyuan Zhuang, Yonghao Zhuang, Joseph~E Gonzalez, et~al.
\newblock Vicuna: An open-source chatbot impressing gpt-4 with 90\%* chatgpt
  quality.
\newblock {\em See https://vicuna. lmsys. org (accessed 14 April 2023)}, 2023.

\bibitem{chowdhery2022palm}
Aakanksha Chowdhery, Sharan Narang, Jacob Devlin, Maarten Bosma, Gaurav Mishra,
  Adam Roberts, Paul Barham, Hyung~Won Chung, Charles Sutton, Sebastian
  Gehrmann, et~al.
\newblock Palm: Scaling language modeling with pathways.
\newblock {\em arXiv preprint arXiv:2204.02311}, 2022.

\bibitem{diao2023lmflow}
Shizhe Diao, Rui Pan, Hanze Dong, Ka~Shun Shum, Jipeng Zhang, Wei Xiong, and
  Tong Zhang.
\newblock Lmflow: An extensible toolkit for finetuning and inference of large
  foundation models.
\newblock {\em arXiv preprint arXiv:2306.12420}, 2023.

\bibitem{ding2021vision}
Henghui Ding, Chang Liu, Suchen Wang, and Xudong Jiang.
\newblock Vision-language transformer and query generation for referring
  segmentation.
\newblock In {\em Proceedings of the IEEE/CVF International Conference on
  Computer Vision}, pages 16321--16330, 2021.

\bibitem{feng2021encoder}
Guang Feng, Zhiwei Hu, Lihe Zhang, and Huchuan Lu.
\newblock Encoder fusion network with co-attention embedding for referring
  image segmentation.
\newblock In {\em Proceedings of the IEEE/CVF Conference on Computer Vision and
  Pattern Recognition}, pages 15506--15515, 2021.

\bibitem{fu2022complexity}
Yao Fu, Hao Peng, Ashish Sabharwal, Peter Clark, and Tushar Khot.
\newblock Complexity-based prompting for multi-step reasoning.
\newblock {\em arXiv preprint arXiv:2210.00720}, 2022.

\bibitem{kazemzadeh2014referitgame}
Sahar Kazemzadeh, Vicente Ordonez, Mark Matten, and Tamara Berg.
\newblock Referitgame: Referring to objects in photographs of natural scenes.
\newblock In {\em Proceedings of the 2014 conference on empirical methods in
  natural language processing (EMNLP)}, pages 787--798, 2014.

\bibitem{kenton2019bert}
Jacob Devlin Ming-Wei~Chang Kenton and Lee~Kristina Toutanova.
\newblock Bert: Pre-training of deep bidirectional transformers for language
  understanding.
\newblock In {\em Proceedings of NAACL-HLT}, pages 4171--4186, 2019.

\bibitem{khot2022decomposed}
Tushar Khot, Harsh Trivedi, Matthew Finlayson, Yao Fu, Kyle Richardson, Peter
  Clark, and Ashish Sabharwal.
\newblock Decomposed prompting: A modular approach for solving complex tasks.
\newblock {\em arXiv preprint arXiv:2210.02406}, 2022.

\bibitem{kirillov2023segment}
Alexander Kirillov, Eric Mintun, Nikhila Ravi, Hanzi Mao, Chloe Rolland, Laura
  Gustafson, Tete Xiao, Spencer Whitehead, Alexander~C Berg, Wan-Yen Lo, et~al.
\newblock Segment anything.
\newblock {\em arXiv preprint arXiv:2304.02643}, 2023.

\bibitem{kojima2022large}
Takeshi Kojima, Shixiang~Shane Gu, Machel Reid, Yutaka Matsuo, and Yusuke
  Iwasawa.
\newblock Large language models are zero-shot reasoners.
\newblock {\em Advances in neural information processing systems},
  35:22199--22213, 2022.

\bibitem{lai2023lisa}
Xin Lai, Zhuotao Tian, Yukang Chen, Yanwei Li, Yuhui Yuan, Shu Liu, and Jiaya
  Jia.
\newblock Lisa: Reasoning segmentation via large language model.
\newblock {\em arXiv preprint arXiv:2308.00692}, 2023.

\bibitem{li2023blip}
Junnan Li, Dongxu Li, Silvio Savarese, and Steven Hoi.
\newblock Blip-2: Bootstrapping language-image pre-training with frozen image
  encoders and large language models.
\newblock {\em arXiv preprint arXiv:2301.12597}, 2023.

\bibitem{liang2023open}
Feng Liang, Bichen Wu, Xiaoliang Dai, Kunpeng Li, Yinan Zhao, Hang Zhang,
  Peizhao Zhang, Peter Vajda, and Diana Marculescu.
\newblock Open-vocabulary semantic segmentation with mask-adapted clip.
\newblock In {\em CVPR}, 2023.

\bibitem{lin2004rouge}
Chin-Yew Lin.
\newblock Rouge: A package for automatic evaluation of summaries.
\newblock In {\em Text summarization branches out}, pages 74--81, 2004.

\bibitem{liu2023gres}
Chang Liu, Henghui Ding, and Xudong Jiang.
\newblock Gres: Generalized referring expression segmentation.
\newblock In {\em CVPR}, 2023.

\bibitem{liu2023visual}
Haotian Liu, Chunyuan Li, Qingyang Wu, and Yong~Jae Lee.
\newblock Visual instruction tuning.
\newblock {\em arXiv preprint arXiv:2304.08485}, 2023.

\bibitem{liu2023llava}
Shilong Liu, Hao Cheng, Haotian Liu, Hao Zhang, Feng Li, Tianhe Ren, Xueyan
  Zou, Jianwei Yang, Hang Su, Jun Zhu, et~al.
\newblock Llava-plus: Learning to use tools for creating multimodal agents.
\newblock {\em arXiv preprint arXiv:2311.05437}, 2023.

\bibitem{liu2023interngpt}
Zhaoyang Liu, Yinan He, Wenhai Wang, Weiyun Wang, Yi Wang, Shoufa Chen,
  Qinglong Zhang, Zeqiang Lai, Yang Yang, Qingyun Li, et~al.
\newblock Interngpt: Solving vision-centric tasks by interacting with chatgpt
  beyond language.
\newblock {\em arXiv preprint arXiv:2305.05662}, 3, 2023.

\bibitem{loshchilov2017decoupled}
Ilya Loshchilov and Frank Hutter.
\newblock Decoupled weight decay regularization.
\newblock {\em arXiv preprint arXiv:1711.05101}, 2017.

\bibitem{luo2020multi}
Gen Luo, Yiyi Zhou, Xiaoshuai Sun, Liujuan Cao, Chenglin Wu, Cheng Deng, and
  Rongrong Ji.
\newblock Multi-task collaborative network for joint referring expression
  comprehension and segmentation.
\newblock In {\em CVPR}, 2020.

\bibitem{mao2016generation}
Junhua Mao, Jonathan Huang, Alexander Toshev, Oana Camburu, Alan~L Yuille, and
  Kevin Murphy.
\newblock Generation and comprehension of unambiguous object descriptions.
\newblock In {\em CVPR}, 2016.

\bibitem{chatGPT}
OpenAI.
\newblock Chatgpt.
\newblock \url{https://openai.com/blog/chatgpt/}.
\newblock Accessed: 2023-09-27.

\bibitem{gpt4v}
OpenAI.
\newblock Gpt-4v(ision) system card.
\newblock \url{https://cdn.openai.com/papers/GPTV_System_Card.pdf}.
\newblock Accessed: 2023-10-09.

\bibitem{ouyang2022training}
Long Ouyang, Jeffrey Wu, Xu Jiang, Diogo Almeida, Carroll Wainwright, Pamela
  Mishkin, Chong Zhang, Sandhini Agarwal, Katarina Slama, Alex Ray, et~al.
\newblock Training language models to follow instructions with human feedback.
\newblock {\em {Adv. Neural Inform. Process. Syst.}}, 35:27730--27744, 2022.

\bibitem{pi2023detgpt}
Renjie Pi, Jiahui Gao, Shizhe Diao, Rui Pan, Hanze Dong, Jipeng Zhang, Lewei
  Yao, Jianhua Han, Hang Xu, and Lingpeng Kong~Tong Zhang.
\newblock Detgpt: Detect what you need via reasoning.
\newblock {\em arXiv preprint arXiv:2305.14167}, 2023.

\bibitem{press2022measuring}
Ofir Press, Muru Zhang, Sewon Min, Ludwig Schmidt, Noah~A Smith, and Mike
  Lewis.
\newblock Measuring and narrowing the compositionality gap in language models.
\newblock {\em arXiv preprint arXiv:2210.03350}, 2022.

\bibitem{ramanathan2023paco}
Vignesh Ramanathan, Anmol Kalia, Vladan Petrovic, Yi Wen, Baixue Zheng, Baishan
  Guo, Rui Wang, Aaron Marquez, Rama Kovvuri, Abhishek Kadian, et~al.
\newblock Paco: Parts and attributes of common objects.
\newblock In {\em Proceedings of the IEEE/CVF Conference on Computer Vision and
  Pattern Recognition}, pages 7141--7151, 2023.

\bibitem{tang2023cotdet}
Jiajin Tang, Ge Zheng, Jingyi Yu, and Sibei Yang.
\newblock Cotdet: Affordance knowledge prompting for task driven object
  detection.
\newblock In {\em Proceedings of the IEEE/CVF International Conference on
  Computer Vision}, pages 3068--3078, 2023.

\bibitem{touvron2023llama}
Hugo Touvron, Thibaut Lavril, Gautier Izacard, Xavier Martinet, Marie-Anne
  Lachaux, Timoth{\'e}e Lacroix, Baptiste Rozi{\`e}re, Naman Goyal, Eric
  Hambro, Faisal Azhar, et~al.
\newblock Llama: Open and efficient foundation language models.
\newblock {\em arXiv preprint arXiv:2302.13971}, 2023.

\bibitem{touvron2023llama2}
Hugo Touvron, Louis Martin, Kevin Stone, Peter Albert, Amjad Almahairi, Yasmine
  Babaei, Nikolay Bashlykov, Soumya Batra, Prajjwal Bhargava, Shruti Bhosale,
  et~al.
\newblock Llama 2: Open foundation and fine-tuned chat models.
\newblock {\em arXiv preprint arXiv:2307.09288}, 2023.

\bibitem{vedantam2015cider}
Ramakrishna Vedantam, C Lawrence~Zitnick, and Devi Parikh.
\newblock Cider: Consensus-based image description evaluation.
\newblock In {\em Proceedings of the IEEE conference on computer vision and
  pattern recognition}, pages 4566--4575, 2015.

\bibitem{wang2023visionllm}
Wenhai Wang, Zhe Chen, Xiaokang Chen, Jiannan Wu, Xizhou Zhu, Gang Zeng, Ping
  Luo, Tong Lu, Jie Zhou, Yu Qiao, et~al.
\newblock Visionllm: Large language model is also an open-ended decoder for
  vision-centric tasks.
\newblock {\em arXiv preprint arXiv:2305.11175}, 2023.

\bibitem{wang2022self}
Xuezhi Wang, Jason Wei, Dale Schuurmans, Quoc Le, Ed Chi, Sharan Narang,
  Aakanksha Chowdhery, and Denny Zhou.
\newblock Self-consistency improves chain of thought reasoning in language
  models.
\newblock {\em arXiv preprint arXiv:2203.11171}, 2022.

\bibitem{wang2022cris}
Zhaoqing Wang, Yu Lu, Qiang Li, Xunqiang Tao, Yandong Guo, Mingming Gong, and
  Tongliang Liu.
\newblock Cris: Clip-driven referring image segmentation.
\newblock In {\em CVPR}, 2022.

\bibitem{wei2022chain}
Jason Wei, Xuezhi Wang, Dale Schuurmans, Maarten Bosma, Fei Xia, Ed Chi, Quoc~V
  Le, Denny Zhou, et~al.
\newblock Chain-of-thought prompting elicits reasoning in large language
  models.
\newblock {\em Advances in Neural Information Processing Systems},
  35:24824--24837, 2022.

\bibitem{yang2022lavt}
Zhao Yang, Jiaqi Wang, Yansong Tang, Kai Chen, Hengshuang Zhao, and Philip~HS
  Torr.
\newblock Lavt: Language-aware vision transformer for referring image
  segmentation.
\newblock In {\em Proceedings of the IEEE/CVF Conference on Computer Vision and
  Pattern Recognition}, pages 18155--18165, 2022.

\bibitem{zhang2023multimodal}
Zhuosheng Zhang, Aston Zhang, Mu Li, Hai Zhao, George Karypis, and Alex Smola.
\newblock Multimodal chain-of-thought reasoning in language models.
\newblock {\em arXiv preprint arXiv:2302.00923}, 2023.

\bibitem{zheng2023ddcot}
Ge Zheng, Bin Yang, Jiajin Tang, Hong-Yu Zhou, and Sibei Yang.
\newblock Ddcot: Duty-distinct chain-of-thought prompting for multimodal
  reasoning in language models.
\newblock {\em arXiv preprint arXiv:2310.16436}, 2023.

\bibitem{zhou2017scene}
Bolei Zhou, Hang Zhao, Xavier Puig, Sanja Fidler, Adela Barriuso, and Antonio
  Torralba.
\newblock Scene parsing through ade20k dataset.
\newblock In {\em Proceedings of the IEEE Conference on Computer Vision and
  Pattern Recognition}, 2017.

\bibitem{zhou2022least}
Denny Zhou, Nathanael Sch{\"a}rli, Le Hou, Jason Wei, Nathan Scales, Xuezhi
  Wang, Dale Schuurmans, Claire Cui, Olivier Bousquet, Quoc Le, et~al.
\newblock Least-to-most prompting enables complex reasoning in large language
  models.
\newblock {\em arXiv preprint arXiv:2205.10625}, 2022.

\bibitem{zhu2023minigpt}
Deyao Zhu, Jun Chen, Xiaoqian Shen, Xiang Li, and Mohamed Elhoseiny.
\newblock Minigpt-4: Enhancing vision-language understanding with advanced
  large language models.
\newblock {\em arXiv preprint arXiv:2304.10592}, 2023.

\bibitem{zhu2023llafs}
Lanyun Zhu, Tianrun Chen, Deyi Ji, Jieping Ye, and Jun Liu.
\newblock Llafs: When large-language models meet few-shot segmentation.
\newblock {\em arXiv preprint arXiv:2311.16926}, 2023.

\bibitem{zou2023generalized}
Xueyan Zou, Zi-Yi Dou, Jianwei Yang, Zhe Gan, Linjie Li, Chunyuan Li, Xiyang
  Dai, Harkirat Behl, Jianfeng Wang, Lu Yuan, et~al.
\newblock Generalized decoding for pixel, image, and language.
\newblock In {\em CVPR}, 2023.

\bibitem{zou2023segment}
Xueyan Zou, Jianwei Yang, Hao Zhang, Feng Li, Linjie Li, Jianfeng Gao, and
  Yong~Jae Lee.
\newblock Segment everything everywhere all at once.
\newblock {\em arXiv:2304.06718}, 2023.

\end{thebibliography}
}

\end{document}